\documentclass[a4paper,10pt]{article}
\usepackage[pdftex]{graphicx}
\usepackage[english]{babel}
\usepackage{biblatex}
\usepackage{arxiv}
\usepackage[utf8]{inputenc} 
\usepackage[T1]{fontenc}    
\usepackage{hyperref}       
\usepackage{url}            
\usepackage{csquotes}
\graphicspath{ {./images/} }

\begin{document}

\title{Large Scale Font Independent Urdu Text Recognition System}
\author{
 Atique Ur Rehman \\
  \texttt{atique.rehman@nu.edu.pk} \\
  \And
 Sibt ul Hussain \\
  \texttt{sibt.ulhussain@nu.edu.pk} \\
}

\maketitle
\begin{abstract}
OCR algorithms have received a significant improvement in performance recently, mainly due to the increase in capabilities of artificial intelligence algorithms. However this advancement is not evenly distributed over all languages. Urdu is among the languages which did not receive much attention, specially in the font independent perspective. There exists no automated system that can reliably recognize printed Urdu text in images and videos across different fonts. To help bridge this gap, we have developed Qaida, a large scale data set with $256$ fonts and complete Urdu lexicon. We have also developed Convolutional Neural Network (CNN) based classification models which can recognise Urdu ligatures with $84.2\%$ accuracy. Moreover we demonstrate that our recognition network can not only recognise the text in the fonts it is trained on, but can also reliably recognise text in unseen (new) fonts. To this end, this paper makes following contributions: (i) we introduce a large scale, multiple (more than $200$) fonts based data set for printed Urdu text recognition \footnote{The data set can be downloaded from \url{https://github.com/AtiqueUrRehman/qaida}}; (ii) we have designed, trained and evaluated a CNN based model for Urdu text recognition; (iii)
we experiment with incremental learning methods to produce state-of-the-art results for Urdu text recognition. All the experiment choices were thoroughly validated via detailed empirical analysis. We believe that our this study can serve as the basis for further improvement in the performance of Urdu OCR systems.
\end{abstract}

\section{Introduction}
Optical Character Recognition (OCR) is the process of converting printed text into computer comprehensible form. Images are not directly comprehensible by computers, rather they are a matrix of numbers with a hidden structure in them. So we need algorithms which can extract those structures and then convert them to a form that can be later indexed, stored and searched by a computer. OCR algorithms are required for detecting and recognizing natural language text from images. OCR algorithms are basis for many advanced applications such as record automation, advanced scanning, cheque verification and advanced data entry methods. Other major applications of OCR include converting legacy literature into editable and searchable form, and its demand is very high for languages with rich literary history, since it will not only help in preserving the literature but will also help visually impaired with advanced reading
devices.\\
OCR algorithms, despite being computationally expensive and difficult to design, have received a significant improvement in performance recently, mainly due to the increase in capabilities of artificial intelligence algorithms. However this advancement is not evenly distributed over all languages. As most advanced OCR systems are designed to work only for Latin based languages with real time performance and accuracy higher than 99\% ~\cite{han2006two}. For instance, OCR systems for English language are even able to read text from natural scene images with same accuracy as of printed text
 ~\cite{jaderberg2016textInTheWild}. However Arabic script based languages have not seen such advancements. Arabic script is a cursive script with many languages including Urdu, Persian and Pushto based upon it. There are many fonts available for Arabic script including Nastaliq, Naskh and Kofi ~\cite{urduFontServer}. A few commercial OCR applications are available
for Arabic script based languages but they lack in performance \cite{tesseractOcr}.\\
Urdu is the super set of Arabic and Persian languages with some additional characters. Due to these additional characters, designing an OCR for Urdu is even more complex than Arabic and Persian. Also any advancement in Urdu OCR will not only be beneficial for Urdu language but will also have its direct impact on Arabic, Persian and other Arabic script based languages. \\
Among the fonts used to print Urdu text, Nastaliq and Naskh are the most common and there have been some successful attempts to build OCR specific to these fonts ~\cite{javed2010segmentation, naz2017urdu}. However there are more then $400$ fonts registered for Urdu ~\cite{urduFontServer} and to the best of our knowledge there does not exist any OCR system that can recognize Urdu text written in all these fonts. One of the major reasons for that is huge inter and intra-class variability found in Urdu characters across fonts. Specifically, Urdu has $39$ letters and is written cursively from right to left. Letters
within a word are joined to form a sub-word called a ligature. Unlike English and other Latin based languages, letters in Urdu are not constrained to only a single shape. Their shape changes with their position within the ligature. Number of shapes per letter vary from one to four. This inter-class variability of Urdu has been one of the main challenges in the development of a robust OCR system. There are almost $18,569$ valid ligatures in Urdu which are to be recognized as compared to only $52$ characters (excluding numbers and
punctuation) in English. Thus designing a font independent Urdu OCR system that can model these intra-class variability becomes the second most important challenge. As mentioned above, Urdu has more than $400$ registered fonts. All these fonts have extreme variations in their writing styles and that also makes it difficult to design an OCR system capable of recognizing Urdu text across all fonts. Figure~\ref{fig:table} illustrates an example ligature in $72$ different fonts. \\
Another important challenge in designing robust OCR is the writing style. As Urdu is commonly written in Nastaliq style, in which scripts are written diagonally with no fixed baseline with may be overlapping ligatures due to lack of any standard. Also, Urdu is a bidirectional language where normal text is written from right to left, while numbers are written from left to right. All these challenges makes it hard to build a robust OCR system and that is the reason why Urdu OCR systems are less mature in performance than other languages.\\
To solve the challenges mentioned above and to capture the strong inter and intra-class variability across available Urdu fonts we have (i) designed a pipeline for synthetically generating textual images of Urdu lexicon in different fonts, (ii) generated a large scale Urdu text recognition data set with complete Urdu Lexicon ($18,569$ ligatures) in $256$ fonts, comprising of more than $4$ million images, and (iii) designed, trained and evaluated a deep Convolutional Neural Network for font independent Urdu text recognition with $86$\% accuracy across complete Urdu lexicon. It is important to mention that our data generation pipeline is language independent and can be used to generate textual images for any other (Arabic script based) language. This data set captures the inter and intra-class variability by including a large number of fonts and any system trained on this data set would be able to recognize Urdu text font independently. Finally our trained CNN can be transfer-learned to achieve superior performance on font specific tasks, is able to recognize new unseen fonts and achieves comparable performance on current benchmark data set named Urdu Printed Text Image Database (UPTI) ~\cite{sabbour2013segmentation}.
Following subsection briefly explains the data set creation steps.

\section{Synthetic Data Generation}
To train a recognition system for detecting Urdu text irrespective of font and size, it is essential to have access to a data set that with multiple fonts, and enough variation in the structure. While there are some publicly available data sets such as UPTI ~\cite{sabbour2013segmentation}, but they lack in Font variations and do not comprise on more then one or two fonts. And this lack of large-scale font-rich recognition data sets have forced previous works to focus on font-specific recognition systems ~\cite{sabbour2013segmentation, nastaliq2013offline, naz2016urdu}. \\
One of the reason for this lack of font-rich data sets is the time consuming process of labeling such a data set. Which is also very expensive in terms of man hours. Our proposed solution to this problem is to synthetically generate a data sets with all the required properties. This would not only avoid all the laborious work required to label the data set, but can also be generalized and applied to similar problems. Similar approach has been successfully used by ~\cite{jaderberg2016textInTheWild} for generating English text-recognition data set, ~\cite{DBLP:journals/corr/GoodfellowBIAS13} for recognizing house numbers from Google street view images and ~\cite{DBLP:journals/corr/JaderbergSVZ14} for text recognition in natural scenes. Following the success of synthetic data in aforementioned scenarios, we propose a synthetic data generation pipeline for font-independent printed Urdu text recognition. Following steps illustrate our proposed pipeline.

\subsection{Font acquisition}
Urdu has two most frequently used fonts, Nastaleeq and Naskh. 
\begin{itemize}
 \item Nastaleeq is a Persian based calligraphic script developed in $14$th century and alongside Urdu it is dominantly used for writing Kashmiri, Punjabi and Pashto languages.
 \item Naskh is more dominant in Arabic language but is also vastly used for Urdu and Pashto. 
\end{itemize}
The popularity of these two fonts have forced most of previous work to focus on either Nastaliq or Naskh font. But these are not the only fonts used in printing Urdu text, there are more then $400$ fonts publicly available on Urdu font web server ~\cite{urduFontServer}. Since our approach focus on making a font independent data set, the first step of our pipeline was to acquire a font database with large number of Urdu fonts. Urdu font web server ~\cite{urduFontServer} is the largest public repository of Urdu fonts. To download all these fonts we developed an automated web scraper. This resulted in a database of approximately $400$ fonts. The next step of the pipeline was to validate that all the acquired fonts support the same unicode set.

\subsection{Font filtering}
Unicode provides full support for Urdu lexicon but there are some problems that are still unresolved ~\cite{ijaz2007Urducorpus}. For example there are some characters with multiple Unicodes, also there are some characters which are connected in nature while the Unicode table provides two different versions of them, one with connected nature and the other with non-connected nature. Due to these discrepancies, different fonts support different Unicode sets. For the next component of our pipeline to work, only one Unicode value per symbol could be supported. So we selected the Unicode set supported by maximum number of fonts and removed all the other fonts. This reduced our font database from $400$ to $256$.

\subsection{Ligature acquisition}
After the selection of fonts the next step was to acquire an Urdu corpus and decide the level of segmentation i.e. word, ligature or character. The level of segmentation is directly related to the methodology used for recognition. Trivial OCR systems work on character level segmentation. The problem with this approach is that during recognition, the word has to be spitted in characters. In case of Latin script based languages, it is a comparatively easier task, since the characters are written separately. But in case of Arabic script based languages, segmentation becomes the bottleneck ~\cite{javed2010segmentation}. The alternate approach is to segment the words on ligature level. There have been some successful attempts for Urdu OCR using ligature based segmentation ~\cite{sabbour2013segmentation, javed2010segmentation} and it also provides a good trade-off between the segmentation complexity in character based approach and the huge vocabulary in word based approach. Based on these observations, we decided to use ligature based segmentation for our pipeline.\\ 
The next step was to acquire a corpus with all possible Urdu ligatures. For this we used a publicly available corpus from Center for Language Engineering (CLE). This corpus have a comprehensive list of Urdu ligatures acquired from over $6$ million lines of text from different sources.

\subsection{Text Rendering}
For Urdu image corpus generation we created a rendering engine that can generate images of all Urdu ligatures in all the selected fonts. There are many existing solutions available in many programming languages, but most of them support only Latin based languages and render images from left to right. Due to the location based connected nature of Urdu if Urdu is rendered from left to right the characters would change their Unicode. So we have to adapt the existing solutions to Arabic script based languages. We developed a text reshaper, that reshapes a ligature and converts it into a reverse form, that can later be rendered using any existing rendering engine.
\subsection{Corpus generation}
\begin{figure}[ht]
 \centering
  \includegraphics[width=\textwidth,]{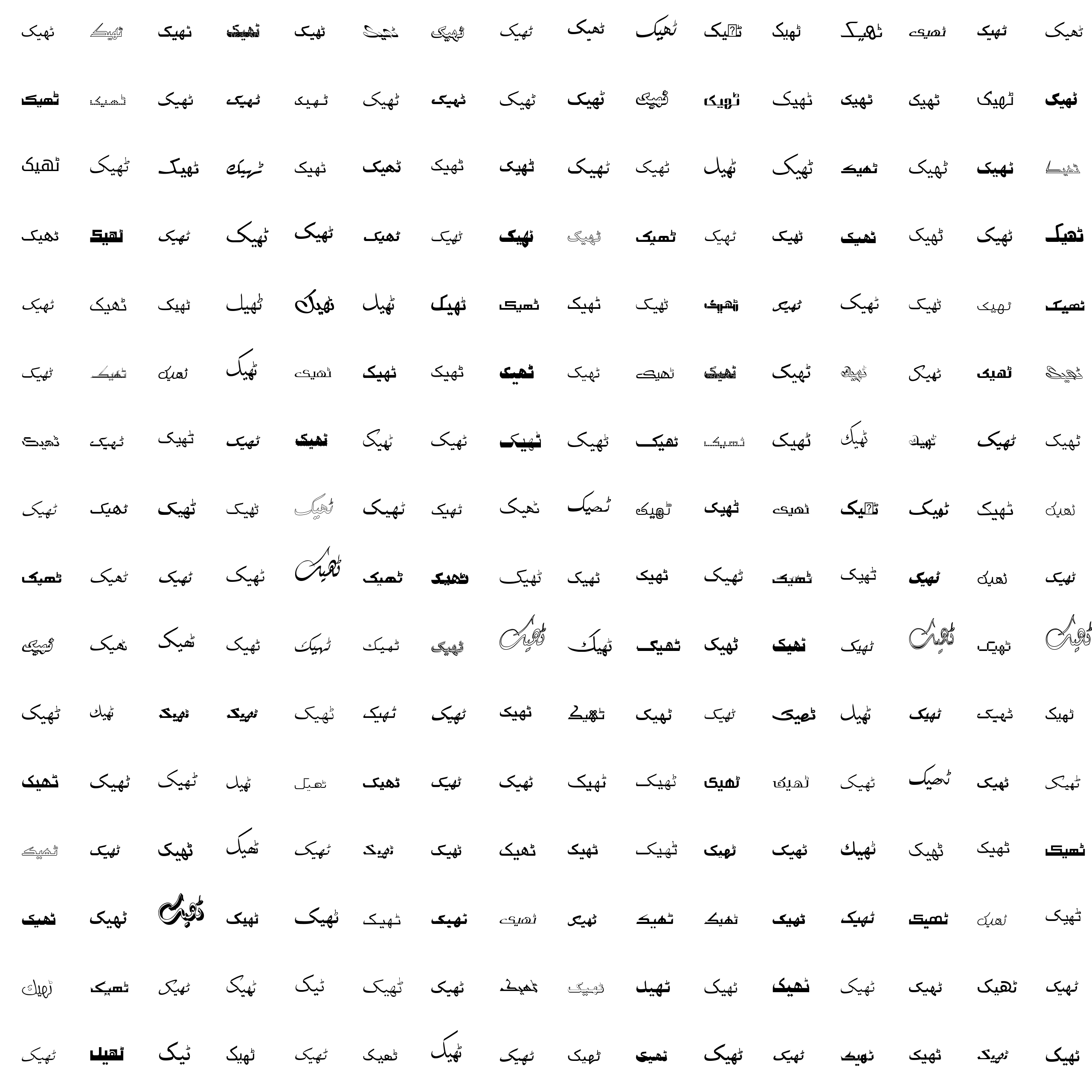}
  \caption{An example ligature generated in $256$ different fonts.}
  \label{fig:table}
\end{figure}

The final phase of the pipeline was to generate the images. A gray-scale image with $160\times160$ pixels was generated for each ligature in all the fonts. Since the total number of ligatures in Urdu language are almost $18565$ and the selected fonts
were $256$, we ended up with a total of almost $47,51872$ images. We also generated some variations of the data set for experimentation, these are as follows:
\begin{itemize}
 \item Since this data set comprises of almost $120$GB a smaller version of the dataset with each image of size $80\times80$ was generated.
 \item A binarized version of the the data set with images thresholded to have only $0$ or $1$ was also generated for evaluations purposes.
 \item Another data set with only $2000$ most occurring ligatures was also generated.
\end{itemize}
We did not augment the data set since it can be handled during training in most of the deep learning libraries ~\cite{tensorflow2015-whitepaper, NEURIPS2019_9015}. Figure \ref{fig:table} illustrates a ligature in $256$ different fonts.

\section{Recognition Algorithm}
Text recognition is an active area of research and there exists many solutions ranging from legacy techniques such as pattern matching \cite{nawaz2009optical} and  Hidden Markov Models (HMM) \cite{javed2013segmentation} to contemporary techniques such as Neural Networks \cite{jaderberg2016textInTheWild}. While methods like HMM have been widely used in recognition tasks, Deep Neural Networks (DNNs) have outperformed all previous techniques on many Computer Vision tasks \cite{krizhevsky2012imagenet, gupta2016synthetic, long2015fully}. Following the success of DNNs on similar tasks, we have designed our system based on DNN.

\subsection{Architecture Selection}
Deep Neural Networks have many types including Convolutional Neural Networks (CNN), Recurrent Neural Networks (RNN) and Fully Convolutional Neural Networks (FCN) etc. All these types have different architectures (arrangement of neurons) and are designed for different tasks. Among them CNNs are mostly used for Computer Vision tasks. The main reason behind the success of CNN on Computer Vision is the Convolutional Layer, which is the core component of a CNN. The Convolutional layer is inspired from the Convolutional operator, widely used in Digital Image Processing. The Convolutional layer has a matrix of learnable parameters, which are convolved over the input to form the output. During the training, these parameters are updated such that they extract useful information from the input at each level. \\
Over the years CNNs have been successfully used in many Computer Vision tasks including Classification ~\cite{krizhevsky2012imagenet} , Localization ~\cite{gupta2016synthetic} and Segmentation ~\cite{long2015fully}. Inspired by the success of CNN on many large-scale classification tasks ~\cite{krizhevsky2012imagenet}, we also designed a CNN based recognition system for Urdu text recognition. 

\subsection{Base Architecture}
There exists a diverse range of architecture variations in CNNs ~\cite{jaderberg2016textInTheWild, gupta2016synthetic, long2015fully, girshick2015fast}, based on the nature of the problem being solved (e.g. Localization, Segmentation or Detection etc.). Architectures also vary on the basis of performance, some are optimized for speed `\cite{redmon2016you} while other are optimized for memory consumption ~\cite{iandola2016squeezenet}. We have chosen our based architecture based on ResNet-18 ~\cite{he2016deep}. The main reasons behind this choice are as follows :
\begin{itemize}
 \item Similar to our systems ResNet is also designed for classification.
 \item It has also been trained on large-scale data set (more then 1 million images).
 \item It has also trained on large ($1000$) number of categories.  
\end{itemize}
ResNet-18 has a total of $18$ layers with the last fully-connected layer having $1000$ neurons for classification on ImageNet ~\cite{krizhevsky2012imagenet} ($1000$ classes). In our case there are $18569$ classes so we modified the last fully-connected classification layer. Further details on the modifications in the network and training process is explained in Section ~\ref{subsec:incrementalLearning}.  

\subsection{Loss Function}
Once we designed our base architecture the next decision was selection of the loss function. ResNet-18 has Softmax Cross Entropy loss function, which has been proved to be very effective in classification tasks \cite{jaderberg2016textInTheWild} so it was our initial choice. But after our initial experiments we faced gradient vanishing problem. And it caused hindrance in network convergence. We have a total of $18569$ classes (The total number of valid ligatures in Urdu language) while the most challenging classification task to date is Imagenet Large-scale Visual Recognition Challenge (ILSVRC) with $1000$ categories. Using a flat N-way classifier in our network was causing the initial probabilities to be very low and  hence the gradients were dieing, slowing down the learning process. Although there have been some attempts to solve this problem but most of the proposed solutions are focused towards Natural Language Processing (NLP) ~\cite{mikolov2011empirical}. Most related solution was \cite{jaderberg2016textInTheWild} based on an incremental leaning procedure, which is explained in the next section.

\subsection{Incremental Learning}
\label{subsec:incrementalLearning}
In practice very few networks are trained from scratch since it requires relatively large data set, huge computation power and it is also harder to train a complex network from scratch e.g. modern CNN architectures requires weeks of training on high-end GPU clusters \cite{szegedy2017inception}. Instead it is a common practice to use a pre-trained network which is trained on a large scale data set such as Imagenet and tune it for the problem at hand. There can be many approaches to use such a model, among them most common are (i) Use the pre-trained CNN as a feature extractor by keeping it's weights fixed and replacing the last fully-connected layer ( which has $1000$ outputs ) with any appropriate classifier, and only
training that classifier, (ii) Fine-tune the CNN by replacing the last fully-connected layer with a randomly initialized layer with required outputs and then training the whole network using back-propagation. The major benefit in fine-tuning instead of training from scratch is that instead of randomly initialized weights, you are starting with weights learned from a large-scale data set and it drastically reduces the training time, in some cases in the magnitude of $10$ or $100$ i.e. the network which was taking three weeks to get to a certain accuracy will now require only a few hours. The constraint on using fine-tuning is that the data-set used for the pre-trained network must be similar to your data set otherwise it will not be very helpful. Since we have $160\times160$ gray-scale images while Imagenet has $256\times256$ RGB images, we cannot use any pre-trained model from Imagenet. Among other pre-trained models, most related was trained on MNIST dataset with $28\times28$ images with $10$ categories, but the small image size and small number of categories makes it also inappropriate to use.\\
As mentioned earlier, the main benefit of fine-tuning or transfer learning is replacing randomly initialized weights with learning weights from a similar data set. Keeping this point in mind we decided to pre-train our own network on a subsample of our data set. For this purpose we selected first $400$ of the easiest classes ( Urdu ligatures consist of 1-8 characters, we sorted them based on the number of characters and selected initial $400$). Choosing this sub-sample will solve both issues (i) the number of classes are very less as compared to complete data set, (ii) it is part of the original data and hense is similar to the complete data. We did all the hyper-parameter tuning and model selection experimentation on this sub-sample and then trained the network to have an acceptable accuracy (~$80$\%). It is worth mentioning that the whole purpose of this exercise was to get some initial weights so we did not include any regularization at this point.  \\
After training the network on $400$ classes, we repeated the step by replacing the last fully-connected layer with $2000$ neurons instead for $400$ and trained for 
$2000$ initial classes. After that we finally replaced the last fully-connected layer with $180569$ neurons and trained on complete dataset.\\
Our final architecture is a Convolutional Neural Network with $18$ Convolutional and $2$ fully-connected layers. The network is inspired by ResNet-18 but we also tried to keep it efficient in terms of total parameters, forward-pass time and total memory
used. L$2$ regularization is used to avoid over-fitting.

\section{Data Splitting Criteria}
The next decision was to come up with an appropriate split for the data to be most beneficial. Usually the splitting criteria used in a random split of $7:2:1$ in training, validation and test set. But in our case this random split can result into a biased sets because this random split will not make sure that each font have same representation in the validation and test set. Also this random split will help us to train out system and then evaluate it's performance on images with same fonts (different geometric variations) but will give us any empirical evidence on new fonts (that are not part of the training set). To solve these issues we came up with the following splitting mechanism:
\begin{itemize}
 \item First we randomly split the data with the ratio of  $75:25$ on the basis of fonts. That means all the images belonging to $56$ randomly selected fonts were separated into an Unseen Test set. This will solve our second problem i.e. evaluating the performance on unseen fonts.
 \item The remaining data ($200$ fonts) was then split with the ratio of $80:10:10$ training, validation and test set.
\end{itemize}
For all the experiments hyper-parameters were tuned using the validation set. While the final results are reported on the test set.
\section{Evaluation Measure}
Our this work is focused on text classification and for classification tasks most commonly used evaluation metrics are F-score and Accuracy ~\cite{naz2016arabic, naz2014optical, ahmad2007urdu}.
\subsection{Accuracy}
Accuracy is the ratio of correctly classified example to total number of examples as shown in \ref{eq_acc}
\begin{equation}
\label{eq_acc}
	{\displaystyle Accuracy = {\frac{correct prediction}{total predictions}}}
\end{equation}
The maximum possible value of Accuracy is $1$ and lowest possible value is $0$. In classification tasks, accuracy is commonly used when all the classes have same number of examples. For cases where the total number of classes are not equal, F-score is preferred.
\subsection{F-score}
F-score can be defined as weighted average of precision and recall. Precision is the ratio of the examples correctly classified into a category to the  total examples
classified into a category, while recall is the ratio of the examples correctly classified into a category to the total examples in that category see eq.~\ref{eq:fScore}.

\begin{equation}
\label{eq:fScore}
	F1 = \frac{2 \cdot precision\cdot recall}{precision+ recall}
\end{equation}

\section{Experiments}
Due to the scale of our data set (~$120$GB, $4$million images) choosing the architecture and tuning all the hyper-parameters on complete data set was not feasible. Hence we started our experiments with initial $400$ classes. We first trained different variants of a ResNet-18 inspired network on these $400$ classes to investigate the effects of pooling, number of Convolution and fully-connected layers, dropout and learning rate. After selecting and training our initial model on $400$ classes we fine-tuned this model to $2000$ and then $18569$ classes. Following sections explain the implementation details of all the stages.

\subsection{Stage-I (400 Classes)}
As explained earlier, we have used an incremental learning methodology to train our network. Hence our initial task was to design and train a DNN that can perform well on first $400$ classes. For this purpose we split the images belonging to first $400$ classes (from training set) into training ($70$\%) and validation ($30$\%). And then tuned our hyper-parameters such as  number of Convolutional, Fully-connected and Pooling layers, number of neurons in fully-connected and Convolutional layers and the placement of different layers in the network. At this stage the main purpose was to design a network that can give a good classification score on our initial $400$ classes by keeping the computation as low as possible. All the weights were initialized using Xavier initialization \cite{pmlr-v9-glorot10a} and the model was trained using Adam optimized \cite{kingma2014adam}. The model achieved a classification accuracy of $90\%$ on the validation set. Since this is not our final model and we have to fine-tune this for $2000$ and then all classes, we did not introduce any type of regularization at this stage. 

\subsection{Stage-II (2000 Classes)}
Once our initial model on $400$ classes was trained, the next step was to train a model for $2000$ classes. Using the transfer learning technique, previously used in Face Recognition \cite{sun2014deep} and Object Detection \cite{shin2016deep} we first created a similar model as the one in Stage-I except the last fully-connected layer.  Last fully-connected layer of our new model had $2000$ neurons instead of $400$ due to increase in the number of classes. Then we initialized all layers except the last fully-connected layer with the weights learned from previous stage. Last layer was initialized with random weights using Xavier initialization \cite{pmlr-v9-glorot10a}. After that the model was trained on $70$\% images belonging to first $2000$ classes of the training set. We also tried some variants of the base model in which some new convolutional and pooling layers were added. Pooling has two significant effects on the performance of a CNN, (i) It reduces the memory footprint since after pooling the feature maps are reduced by the stride size. (ii) It makes the model invariant to small shifts and distortions in input, since it merge semantically similar features \cite{lecun2015deep}. Our final model on $2000$ classes achieved $89\%$ accuracy on the validation set.\\

\subsection{Stage-III (All Classes)}
Final stage was to train the model on the complete data set with $18569$ classes. Similar to Stage-II we initialized all layers except the last fully-connected layer
with the weights learned from Stage-II. After that this model was trained on complete training set. During training the data was augmented with geometric variations. 
While training deep learning models using Adam, appropriate scheduling of learning rate can lead to faster convergence and better performance \cite{senior2013empirical}. It is a common practice to reduce the learning rate by half, when the learning curve becomes plateau \cite{zhang2017shufflenet}, we have also used similar approach while training where we started with a learning rate of $0.001$ and reduced it with a factor of $0.5$ at each plateau. After completion the final accuracy on validation set was $0.853$.

\subsection{Font Specific fine-tining}
One of the characteristic of a DNN trained on a large-scale data set is its capability to be fine-tuned to similar data sets, this approach has been successfully used in
fine-tuning models trained on Imagenet \cite{naz2016urdu} on tasks like Semantic Segmentation \cite{long2015fully} and object detection \cite{ren2015faster}. This approach has been proved to be very useful specially when the amount of labeled data is limited \cite{lecun2015deep}. In case of fine-tuning the convergence is also faster as compared to training the same model from scratch.\\ 
In our next experiment we tried to evaluate the performance of fine-tuning our trained model to one of the unseen fonts. This can be very useful for scenarios in which the task is to get best possible performance on a specific font instead of generalizing on all fonts. As mentioned in data splitting section, we have kept an Unseen set containing images generated in $56$ fonts, which are not used in training or validation. For fine-tuning we have chosen a font from this set. The images 
belonging to this font were split into training ($70$\%) and test ($30$\%) set. We trained the model with the learning rate $0.00005$ for $5$ epochs and the model was able to achieve accuracy of $95.01\%$. The same model when trained from scratch was not able to achieve same performance even after $15$ epochs. This shows that the model has the capabilities to be used for similar tasks and can also achieve better performance if the target is a specific subset of fonts.

\section{Conclusion}
In this paper we have presented a large-scale multi-font printed Urdu text recognition data set, comprising of $4$ million images of complete Urdu lexicon in $256$ fonts. The data set has enough inter and intra-class variation to be used for font independent Urdu text recognition. We have also trained ResNet-18 inspired CNN on complete data set with a final accuracy of $85\%$. Finally the models can be fine-tuned on a specific font to achieve font-specific superior performance. The data set is publicly released\footnote{The data set can be downloaded from \url{https://github.com/AtiqueUrRehman/qaida}}, along with the trained models for further research.
\printbibliography
\end{document}